\documentclass{article}
\usepackage{spconf,amsmath,graphicx}
\usepackage{epstopdf}
\usepackage{overpic}
\usepackage{multirow}
\usepackage{hyperref}
\usepackage{algorithm}
\usepackage{algorithmic}
\usepackage{amssymb}
\usepackage{multirow}
\usepackage{booktabs}

\title{Layer-wise Customized Weak Segmentation block and AIoU loss for accurate Object Detection}

\name{Keyang Wang \qquad Lei Zhang$^{\star}$ \qquad Wenli Song \qquad Qinghai Lang \qquad  Lingyun Qin}
\address{School of Microelectronics and Communication Engineering, Chongqing University, Chongqing}

\begin{document}
%
\maketitle
\begin{abstract}
The anchor-based detectors handle the problem of scale variation by building the feature pyramid and directly setting different scales of anchors on each cell in different layers. However, it is difficult for box-wise anchors to guide the adaptive learning of scale-specific features in each layer because there is no one-to-one correspondence between box-wise anchors and pixel-level features. In order to alleviate the problem, in this paper, we propose a scale-customized weak segmentation (SCWS) block at the pixel level for scale customized object feature learning in each layer. By integrating the SCWS blocks into the single-shot detector, a scale-aware object detector (SCOD) is constructed to detect objects of different sizes naturally and accurately. Furthermore, the standard location loss neglects the fact that the hard and easy samples may be seriously imbalanced. A forthcoming problem is that it is unable to get more accurate bounding boxes due to the imbalance. To address this problem, an adaptive IoU (AIoU) loss via a simple yet effective squeeze operation is specified in our SCOD. Extensive experiments on PASCAL VOC and MS COCO demonstrate the superiority of our SCOD.
\end{abstract}
\section{Introduction}
Recently, object detection, as one of the core tasks in computer vision, has achieved significant advances in recent years. On the whole, all the detectors that use deep CNN can be divided into two categories: two-stage approaches \cite{girshick2015fast, ren2015faster, lin2017feature,dai2016r, he2017mask, cai2018cascade, Cao_2020_CVPR, pang2019libra,li2019scale} and one-stage approaches \cite{liu2016ssd, fu2017dssd, redmon2016you, lin2017focal, zhang2018single, zhao2019m2det, redmon2016you,wang2020single, wu2019iou, Zhu_2019_CVPR}. The two-stage detectors have achieved the top performance on benchmark datasets, while the one-stage approaches achieve time efficiency.

It is well known that the problem of scale variation across object instances has a great influence on the object detection task. Most of the anchor based detectors, such as SSD \cite{liu2016ssd}, FPN \cite{lin2017feature}, handle this problem by building the feature pyramid and directly setting different scales of anchors on each feature map cell in multiple feature layers, which makes each level feature map perform its duty. The box-wise anchors are used to guide the learning of scale-specific features at each layer. However, as the feature of each cell is not equal to the box-wise anchor, the scale-specific features of each layer can not always be obtained under the guidance of box-wise anchors due to the lack of adaptive guidance. That is, the anchor based detectors lack of scale-awareness, which is harmful for accurately detecting objects of different scales.

\begin{figure}
\centering
\includegraphics[width=1.0\linewidth]{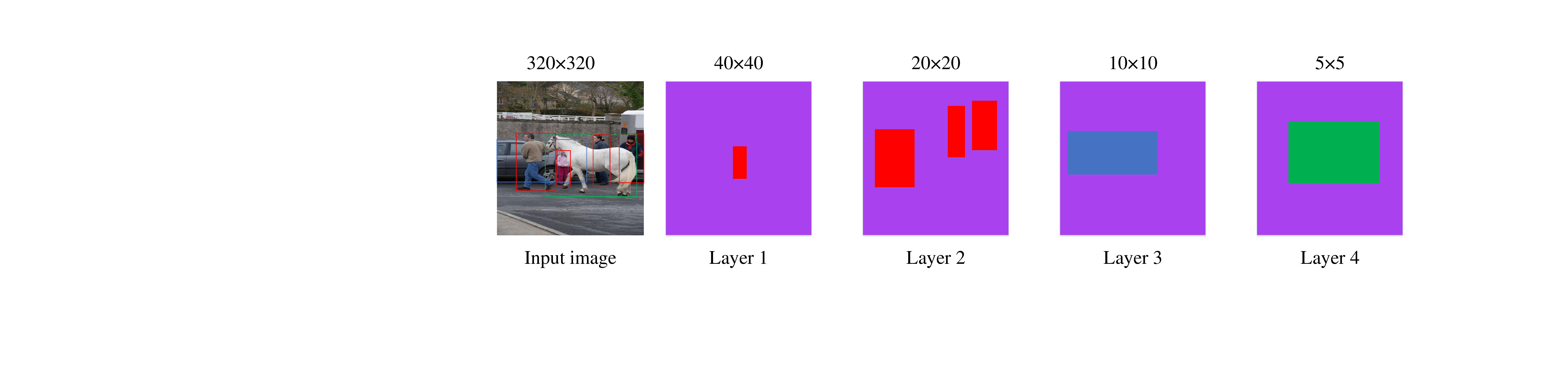}
\vspace{-0.9cm}
\caption{An example of scale-customized weak segmentation idea. The first one is input image which includes six object instances at four different scale range. The remaining four are the results of the scale-customized weak segmentation. All the object instances are assigned to different layers at pixel level according to their scales.}
\vspace{-0.5cm}
\label{example}
\end{figure}

Since the scale-specific features can not be obtained under the guidance of box-wise anchors, we propose an idea to construct an adaptive layer-wise guidance module at the pixel level instead of box level to materialize the scale of each layered features. Motivated by this, in this paper, we introduce a scale-customized weak segmentation (SCWS) block, which is supervised by weakly segmented ground-truth (coarse annotation). The SCWS blocks specify the role of each layer at pixel level according to scale and assist each layer in learning scale-specific features. With the SCWS blocks, the detector will naturally realize the scale-awareness. As shown in Fig.~\ref{example}, we present a picture with six object instances of four different scales. Take the smallest scale (i.e., the little girl) as an example, this bounding box of the little girl is assigned to the lowest layer (layer 1). All pixels falling into the bounding box of the little girl will be uniquely classified as ``person'' category in layer 1, while in other layers, the pixels in the location of the girl will be classified as ``background''. Similarly, each object instance can be assigned to a certain layer at pixel level.

\begin{figure}
 \centering
 \begin{overpic}[width=0.70\linewidth]{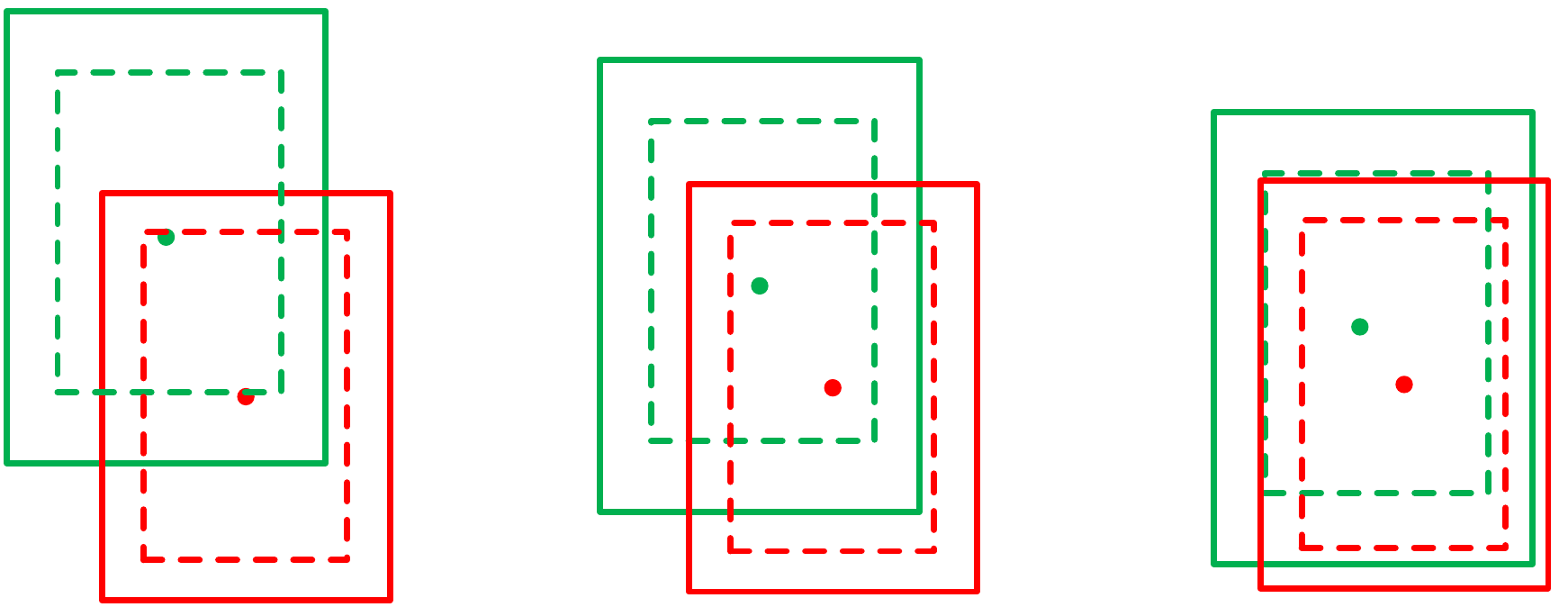}
 \put(2,-4){\scriptsize $\mathcal{L}_{IoU}=0.72$}
 \put(1,-8.5){\scriptsize $\mathcal{L}_{AIoU}=0.80$}
 \put(0,-14){\scriptsize $\Delta^{'}\mathcal{L}_{AIoU}=0.11$}
 \put(39,-4){\scriptsize $\mathcal{L}_{IoU}=0.61$}
 \put(38,-8.5){\scriptsize $\mathcal{L}_{AIoU}=0.69$}
 \put(37,-14){\scriptsize $\Delta^{'}\mathcal{L}_{AIoU}=0.13$}
 \put(75,-4){\scriptsize $\mathcal{L}_{IoU}=0.42$}
 \put(74,-8.5){\scriptsize $\mathcal{L}_{AIoU}=0.49$}
 \put(73,-14){\scriptsize $\Delta^{'}\mathcal{L}_{AIoU}=0.17$}
 \end{overpic}
 \vspace{0.6cm}
 \caption{Examples of the AIoU loss. The red and green boxes (solid) denote the ground-truth (GT) box and predicted bbox, respectively. The red and green bboxes (dashed) denote the squeezed (scaled) GT bbox and predicted bbox, respectively.}
 \vspace{-0.4cm}
  \label{pious}
\end{figure}

Furthermore, the matching strategy based on IoU is used for preparing the samples in detection tasks. However, this strategy neglects a fact that the IoU distribution of all the samples is imbalanced. Generally, the number of samples at low IoU levels (i.e. hard samples) is significantly larger than that at high IoU levels (i.e. easy samples). Needless to say, the hard samples dominate the gradients of the location loss during training, which enables the model to be biased toward the hard samples.
In order to alleviate the model bias, we propose an adaptive IoU (AIoU) via a simple yet effective squeeze operation, as shown in Fig.~\ref{pious}. And based on the AIoU, we propose a novel AIoU loss, which can essentially improve the hard samples dominated imbalance problem. There are two advantages of our AIoU loss over the standard IoU loss. First, the $\mathcal{L}_{AIoU} \ge \mathcal{L}_{IoU}$, which means our AIoU loss is stricter than standard IoU loss for localization. Second, the relative change rate of AIoU loss to the standard IoU loss, i.e., $\Delta^{'}\mathcal{L}_{AIoU}$, increases with the increase of IoU value. This means that our $\mathcal{L}_{AIoU}$ pays more attention to the samples at high IoU levels, which can prevent the network from being biased toward the hard samples.

\begin{figure*}
\centering
\includegraphics[width=0.67\linewidth,height=6cm]{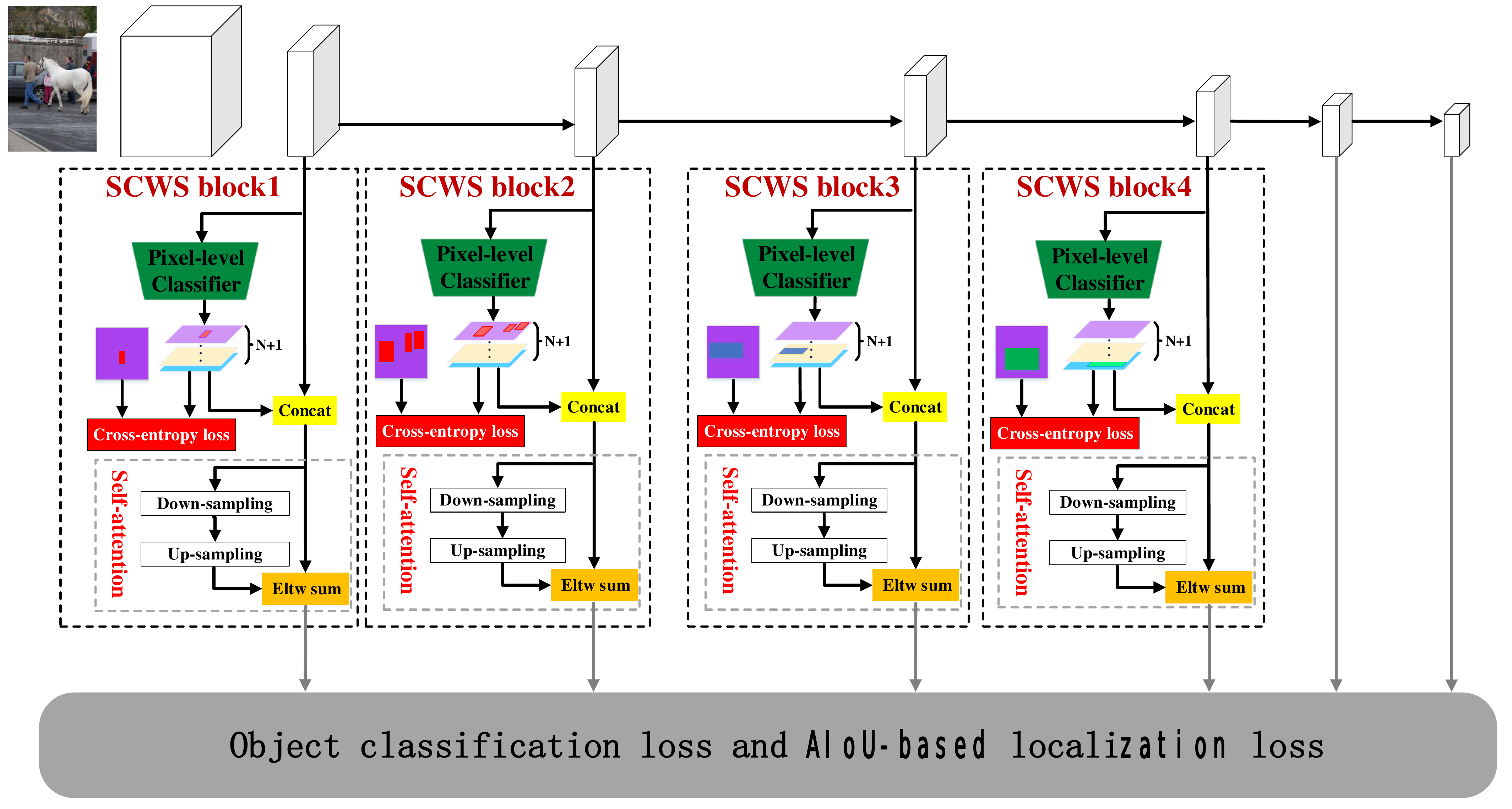}
\vspace{-0.4cm}
\caption{The overall architecture of SCOD (320$\times$320) which includes four SCWS blocks. Each block includes a coarse label guided pixel-level classifier for weak segmentation and an easily operated self-attention unit for feature enhancement.}
\label{sgwsNet}
\vspace{-0.5cm}
\end{figure*}

With the SCWS blocks for layer-wise scale-specific feature learning and the AIoU loss for training, a scale-aware object detector (SCOD) is finally proposed in this paper. In summary, this paper makes three contributions:
\textbf{(1)} We propose a pixel-level scale-customized weak segmentation (SCWS) idea to learn scale-specific feature in each layer, which can make the detector attentively realize scale-awareness.
\textbf{(2)} We introduce an adaptive IoU (AIoU) loss to prevent the gradients of localization loss from being dominated by outliers and ensure accurate bounding box regression of the detector.
\textbf{(3)} With multiple SCWS blocks and the AIoU loss, a scale-aware object detector (SCOD) is materialized in this paper.
\vspace{-0.1cm}

\section{The Proposed Approach}
\vspace{-0.1cm}
\subsection{Principle of SCWS Blocks}
\vspace{-0.2cm}
We illustrate the network structure of our SCOD in Fig.~\ref{sgwsNet}, in which four proposed scale-customized weak segmentation (SCWS) blocks and the adaptive IoU (AIoU) loss are deployed for single-shot detection.

\textbf{Basic SCWS block}. As shown in Fig.~\ref{sgwsNet}, we embed a SCWS block in each of the four different layers to learn scale-specific features. In each SCWS block, a scale-aware pixel-level classifier is added at each SCWS block to implement scale-customized weak semantic segmentation. A 3 $\times$ 3 convolutional filter is applied as the pixel-level classifier. The classifier has to achieve the following two goals. (1) Classify each pixel in each layer. (2) Determine whether the current pixel belongs to the object instances of the customized layer and assist the learning of scale-specific feature.

In order to achieve the two goals described above, the first problem to be solved is how to generate scale-specific weak segmentation ground-truth for training this scale-aware pixel-level classifier. General segmentation algorithms require pixel-level image annotation, but this is not feasible in the object detection task. Instead, we take the bounding-box level annotations of detection task as weak segmentation labels to perform weakly supervision learning of the segmenter.

Let $F_i \in \mathbb{R}^{H \times W \times C}$ be the feature map at layer $i$ of a backbone. For each pixel $(x,y)$ on the feature map $F_i$, we can map it back to the input images as $(x^*,y^*)$. If the pixel $(x^*,y^*)$ falls into any GT bbox and simultaneously the scale of this GT bbox falls within the customized range in layer $i$, we then assign the category label of the GT bbox to pixel $(x,y)$. If it is located in multiple bounding boxes which meet the above requirements, the label of the ground-truth with the smallest size is selected. If it does not locate in any ground-truth or that the scale of this ground-truth bbox does not fall within the customized range of the layer $i$, we will assign it to the background class. Mathematically, if we set the scale range of the layer $i$ as ($s_i^l$, $s_i^h$), the label $G_{xy}$ of the pixel $(x,y)$ for weak segmentation can be defined as follows:
\vspace{-0.2cm}
\begin{equation}
\vspace{-0.2cm}
\label{eq2}
G_{xy}=\left\{
\begin{aligned}
class \quad c, \ \ &if\ the \ pixel\ (x,y) \ falls \ into  \\
&\hspace{0.1cm}\ a \ GT \ and \ S_b \in (s_i^l, s_i^h) \\
class \quad 0, \ \ & \quad \quad \quad else
\end{aligned}
\right.
\vspace{-0.1cm}
\end{equation}
where the $c \in \{1,2, \cdots, N \}$ is the category label of the GT, $N$ is the number of classes, and $S_b$ is the scale of the matching ground-truth box. The area is adopted as the scale standard.

In the following, we further elaborate how to customize the scale range ($s_i^l$, $s_i^h$) of each feature layer. As the proposed SCOD is an anchor based detector, we therefore propose to customize the scale range based on the scale of anchors. For example, if the scale range of the anchors is set as ($a_{min},a_{max}$) at layer $i$, then the scale range ($s_i^l$, $s_i^h$) in Eq.(\ref{eq2}) is specified as: $s_i^l=s_{i-1}^h$, $s_i^h=(a_{min}a_{max}+a_{max}^2)/2$. In particular, we set the $s_1^l=0$ in this paper.

After generating the weak segmentation ground-truth labels, the scale-aware pixel-level classifier can be trained to achieve the two goals described above. As is shown in Fig.~\ref{sgwsNet}, we slide the convolutional filter to compute the prediction $Y \in \mathbb{R}^{(N+1) \times H \times W}$ of the segmenter as follows:
\vspace{-0.2cm}
\begin{small}
\begin{equation}
\label{eq4}
\vspace{-0.2cm}
Y=\mathcal {F}{(F_i)}
\vspace{-0.1cm}
\end{equation}
\end{small}
where $\mathcal {F}$ is a classification function. The $Y$ is a classification score vector that satisfies the following requirements:
\vspace{-0.2cm}
\begin{small}
\begin{equation}
\label{eq5}
\vspace{-0.2cm}
Y \in [0,1]^{(N+1)\times H \times W} \quad and \quad \sum\limits_{c=0}^{N}Y_{c,x,y}=1
\vspace{-0.1cm}
\end{equation}
\end{small}
Then an extra pixel-level cross-entropy loss function $\mathcal{L}_{scws}$, defined as Eq.(\ref{eq4}), is adopt to optimize this classifier and induce the learning of scale-specific features.
\vspace{-0.2cm}
\begin{small}
\begin{equation}
\vspace{-0.2cm}
\label{eq4}
\mathcal{L}_{scws}(I,G)=-\frac{1}{HW}\sum\limits_{x=1}^W\sum\limits_{y=1}^H log(Y_{G_{xy},x,y})
\vspace{-0.1cm}
\end{equation}
\end{small}
where $I$ is the image, $Y \in \mathbb{R}^{(N+1) \times H \times W}$ is the output of the pixel-level classifier, $G \in \{0,1,2,\cdots,N\}^{H \times W} $ is the human-intervened label generated by bounding-box annotation, and $N$ is the number of classes excluding background.

After obtaining the weak segmentation results, they are concatenated with the corresponding convolution feature maps as scale-specific features, as shown in Fig.~\ref{sgwsNet}, which can also enrich the semantic and detailed information of different layered output. In order to further enhance the features of each layer, we add an often used self-attention unit at each layer, which includes a downsampling and an upsampling.
\vspace{-0.2cm}
\subsection{AIoU loss}
\vspace{-0.2cm}
Intersection over Union (IoU) is a popular metric in object detection, defined as:
\vspace{-0.2cm}
\begin{small}
\begin{equation}
\vspace{-0.2cm}
\label{eq5}
IoU=\frac{|B_p \cap B_g|}{|B_p \cup B_g|}
\vspace{-0.0cm}
\end{equation}
\end{small}
where $B_p$ is the predicted bbox, $B_g$ is the ground-truth. As suggested in \cite{yu2016unitbox}, the IoU loss, defined as Eq.(\ref{eq6}), is more suitable to obtain the optimal IoU metric and regress accurate bounding boxes than $\mathcal{L}_n$-norm loss function.
\vspace{-0.2cm}
\begin{equation}
\label{eq6}
\mathcal{L}_{IoU}=1-IoU
\vspace{-0.2cm}
\end{equation}
However, the gradient of the $\mathcal{L}_{IoU}$ is the constant 1 for both easy samples (samples at high IoU levels) and hard samples (samples at low IoU levels). Needless to say, the hard samples dominate the gradients of the localization loss in the training phase as the number of samples at low IoU levels is larger than that at high IoU levels. This enables the detection model to be biased toward the hard samples. In order to mitigate the adverse effects, on the basis of the IoU, we propose an adaptive extension of the IoU, named AIoU. As shown in Fig.~\ref{pious}, we first compress both the predicted bboxes $B_p$ and the corresponding ground-truth $B_g$ at the same scale ratio $\beta$ to obtain the squeezed prediction bbox $B_p^c$ and the corresponding squeezed ground-truth $B_g^c$. Then we calculate the IoU between the compressed bboxes to obtain the AIoU. After obtaining AIoU, similar to IoU, the AIoU loss is defined as:
\vspace{-0.2cm}
\begin{equation}
\vspace{-0.2cm}
\label{eq7}
\mathcal{L}_{AIoU}=1-AIoU
\vspace{-0.2cm}
\end{equation}

\begin{figure}
\centering
\includegraphics[width=0.8\linewidth]{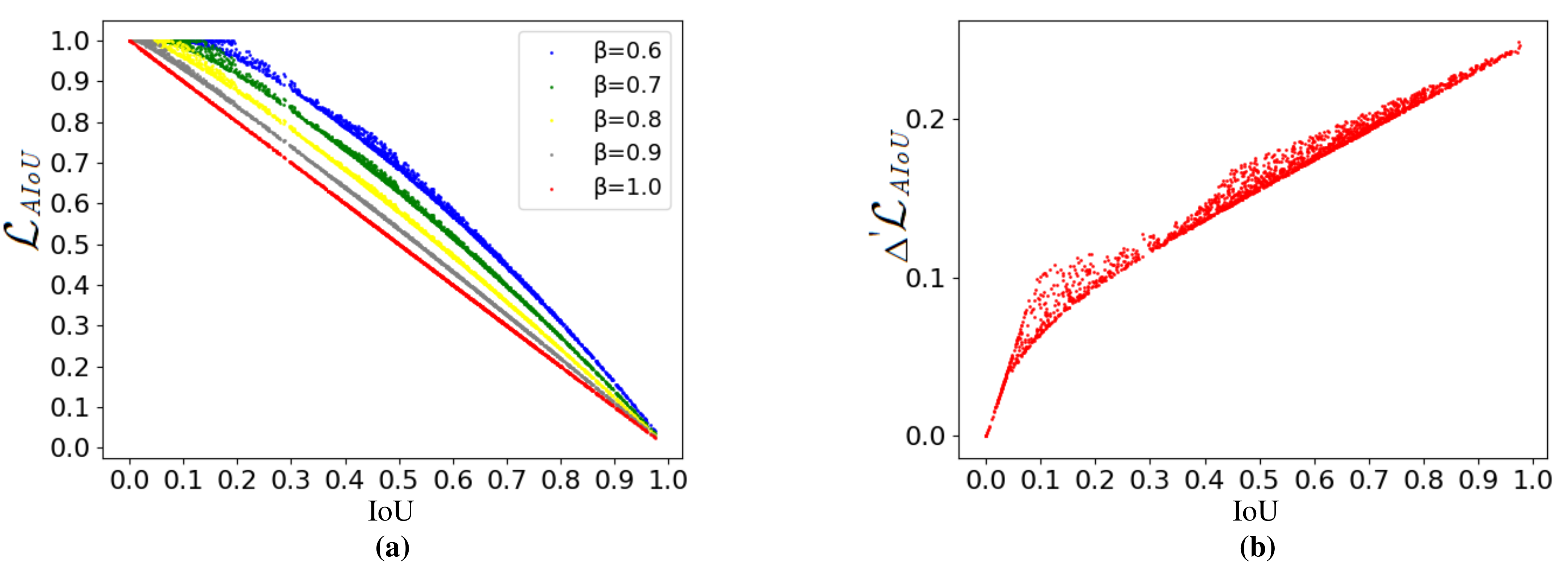}
\vspace{-0.5cm}
\caption{Two distributions of $\mathcal{L}_{AIoU}$ of 2000 bounding boxes. (a) is the distribution of $\mathcal{L}_{AIoU}$ at different scale ratio $\beta$. (b) is the distribution of $\Delta^{'}\mathcal{L}_{AIoU}$ described in Eq.(\ref{eq8}).}
\label{piouloss}
\vspace{-0.5cm}
\end{figure}
\textbf{In-depth Analysis and Comparison to standard IoU loss}. Compared with the standard IoU loss, our proposed AIoU loss can balance the hard samples and easy samples and prevent the gradients of localization loss from being dominated by hard samples. We will give a in-depth analysis from two aspects. Firstly, we visualize the distribution of $\mathcal{L}_{AIoU}$ and $\mathcal{L}_{IoU}(\beta=1.0)$ of 2000 bounding boxes in Fig.~\ref{piouloss} (a). We can find that the distribution of $\mathcal{L}_{AIoU}$ shows a convex shape with the increase of IoU, which means AIoU loss can increase the gradients of samples at high IoU levels to a certain extent and avoid the whole gradients being completely dominated by the hard samples. Secondly, we define the relative change rate of AIoU loss to the IoU loss as follows:
\vspace{-0.1cm}
\begin{equation}
\label{eq8}
\Delta^{'}\mathcal{L}_{AIoU}=\frac{\mathcal{L}_{AIoU}-\mathcal{L}_{IoU}}{\mathcal{L}_{IoU}}
\vspace{-0.2cm}
\end{equation}
We also visualize the distribution of $\Delta^{'}\mathcal{L}_{AIoU}$ in Fig.~\ref{piouloss} (b) while the scale ratio is 0.8. We can clearly find that the $\Delta^{'}\mathcal{L}_{AIoU}$ increases with the increase of IoU, which means our AIoU loss can automatically increase the weight of the easy samples compared with standard IoU loss. Training with AIoU loss, the huge amount of cumulated gradient produced by easy examples can be up-weighted and the outliers can be relatively down-weighted, which can prevent the network from being biased toward the hard samples.

With the proposed segmentation loss, AIoU loss and the basic object classification loss, the overall objective of our SCOD is summarized as:
\begin{small}
\begin{equation}
\vspace{-0.3cm}
\label{eq11}
\mathcal{L}(I,B_g,G)=\mathcal{L}_{cls}(I,B_g)+\mathcal{L}_{scws}(I,G)+\mathcal{L}_{AIoU}(I,B_g)
\vspace{-0.4cm}
\end{equation}
\end{small}

\section{Experiments}
\vspace{-0.3cm}
In this section, we first conduct experiments on two benchmarks for the object detection task, i.e., PASCAL VOC \cite{everingham2007pascal} dataset and MS COCO \cite{lin2014microsoft} dataset. We then conduct ablation analysis of the proposed SCWS blocks and AIoU loss in our SCOD. All of our models are trained under the PyTorch framework with SGD solver on NVIDIA Titan Xp GPUs.
\vspace{-0.4cm}
\subsection{Experiments on PASCAL VOC}
\vspace{-0.2cm}
In our experiments, all models are trained on the union of the VOC 2007 \texttt{trainval} and VOC 2012 \texttt{trainval} datasets, and tested on the VOC 2007 \texttt{test} set. A stricter COCO-style Average Precision (averaged AP at IoUs from 0.5 to 0.9 with an interval of 0.1) metrics is adopted on the VOC dataset.

We compare our method with the state-of-the-art detectors in Table~\ref{sample-table1}. Our detector can improve AP by 2.9$\%$ compared with the baselines DSSD320 (from 56.1$\%$ to 59.0$\%$) when the input size is 320. Particularly, the improvement for AP at higher IoU threshold (0.8, 0.9) is about 4$\%$ over DSSD, which demonstrates that our method can substantially improve the model localization accuracy. And the AP of our detector (59.0$\%$) is better than most of state-of-the-art detectors on PASCAL VOC, such as the RefineDet320 (54.7$\%$), DAFS (58.7$\%$), Cascade R-CNN (58.5$\%$). When the input size is increased to 512, our method can achieve 60.0$\%$ AP which is comparable to most of SOTA detectors at the same scale.
\begin{table}[h]
\vspace{-0.5cm}
  \caption{Comparison of methods on the PASCAL VOC.}
  \label{sample-table1}
  \centering
  \footnotesize
  \resizebox{8.2cm}{2.6cm}{
  \begin{tabular}{c|c|p{0.5cm}p{0.5cm}p{0.5cm}p{0.5cm}}
    \toprule
    Method      &     Backbone       &   AP& AP$_{50}$ & AP$_{70}$  & AP$_{90}$   \\
    \hline
    \footnotesize\emph{two-stage:}& & &&&\\
    Faster R-CNN \cite{ren2015faster}              & ResNet-50-FPN   &52.9&79.8&61.7&8.8  \\
    Cascade R-CNN \cite{cai2018cascade}            & ResNet-50-FPN   &58.5&80.0 &65.8& 21.5    \\
    \hline
    \hline
    \footnotesize\emph{one-stage:}& & &&&\\
    SSD300 \cite{liu2016ssd}             & VGG-16  &52.7&77.6 &61.0&11.4  \\
    DSSD320 \cite{fu2017dssd}             & ResNet-50     &56.1&79.6 &64.1&16.0 \\
    RefineDet320 \cite{zhang2018single} & VGG-16           &54.7&80.0 &63.5& 12.2\\
    DAFS320 \cite{li2019dynamic}              & ResNet-101       &58.7 &\textbf{81.0} & 66.9  &20.0 \\
    \textbf{SCOD320}              & ResNet-50  & \textbf{59.0}&80.5&\textbf{67.2}&\textbf{20.2} \\
    \hline
    SSD512 \cite{liu2016ssd}              & VGG-16      &57.5&79.8&66.7 &15.2 \\
    DSSD512 \cite{fu2017dssd}              & ResNet-50       &58.5&81.5&67.6 &15.8\\
    RefineDet512 \cite{zhang2018single} & VGG-16            &58.4&81.8& 67.2& 15.6            \\
    RetinaNet \cite{lin2017focal} & ResNet-101-FPN           &59.3&81.1&67.5  &20.1            \\
    DAFS512 \cite{li2019dynamic}              & VGG-16       &59.4&\textbf{82.4}&67.6 &18.0 \\
    \textbf{SCOD512 }           & ResNet-50      &\textbf{60.0} &82.1&\textbf{68.2}&\textbf{20.2}   \\
    \bottomrule
  \end{tabular}
  }
  \vspace{-0.4cm}
\end{table}
\vspace{-0.4cm}
\subsection{Ablation Study}
\vspace{-0.3cm}
We conduct the ablation experiments on PASCAL VOC. We analyze the effectiveness of SCWS blocks and AIoU loss.

\textbf{Scale-customized weak segmentation (SCWS) block}. We first conduct ablation experiments to verify the effectiveness of each SCWS block. As show in Table~\ref{sample-table2}, we can find that each SCWS block makes different contribution to the overall detector. And we get the best AP (57.7$\%$) when all four SCWS blocks are added to our detector, which outperforms the baseline DSSD320 by 1.6$\%$ AP (from 56.1$\%$ to 57.7$\%$). This means that our SCWS blocks can really induce each layer in learning scale-specific features.
\begin{table}[h]
\vspace{-0.5cm}
  \caption{Ablation results of each component (i.e., SCWS block and AIoU loss) in SCOD.}
  \footnotesize
  \label{sample-table2}
  \centering
  \begin{tabular}{ccp{0.35cm}p{0.35cm}p{0.35cm}p{0.35cm}}
    \toprule
    \multicolumn{2}{c}{Method }              &  AP& AP$_{50}$  & AP$_{70}$  & AP$_{90}$   \\
    \hline
     \multicolumn{2}{c}{Baseline(DSSD) }      &56.1 & 79.6  &64.1&16.0   \\
     \multicolumn{2}{c}{Baseline+SCWS block1}   &56.5&80.0&64.7&16.4\\
     \multicolumn{2}{c}{Baseline+SCWS block1,2}   &57.1&80.4&65.3&17.0\\
     \multicolumn{2}{c}{Baseline+SCWS block1,2,3}   &57.5&80.5&66.0&17.6\\
     \multicolumn{2}{c}{Baseline+SCWS block1,2,3,4}   &57.7&\textbf{80.7}&66.2&17.9\\
     \multicolumn{2}{c}{Baseline+SCWS block1,2,3,4+AIoU loss }  & \textbf{59.0}&80.5&\textbf{67.2}&\textbf{20.2}\\
    \bottomrule
  \end{tabular}
  \vspace{-0.3cm}
\end{table}
\begin{table}[h]
\vspace{-0.3cm}
  \caption{Performance analysis of the squeezed ratio $\beta$.}
  \scriptsize
  \label{sample-table4}
  \centering
  \begin{tabular}{p{0.8cm}p{0.8cm}p{0.9cm}p{0.9cm}p{0.9cm}p{0.9cm}}
    \toprule

     \ $\beta$               & \ 1.0&\ 0.9 &\ 0.8 &\ 0.7 &\ 0.6   \\
    \hline
    AP              & 57.9 & 58.6 & \textbf{59.0} &58.6  & 58.0   \\
    \bottomrule
  \end{tabular}
  \vspace{-0.5cm}
\end{table}

\textbf{Performance analysis of our AIoU loss}. We conduct another two experiments to validate the effectiveness of the AIoU loss. Firstly, we use the AIoU loss instead of Smooth-L1 loss to train our SOCD. As shown in Table~\ref{sample-table2}, The detector trained with AIoU loss can improve AP by 1.3$\%$ compared with the detector trained with Smooth-L1 loss. Secondly, we conduct experiments on our SCOD to analyze the effect of the squeezed ratio $\beta$. Results are presented in Table~\ref{sample-table4}, in which five different ratio ranging from 0.6 to 1.0 are experimented. Note that $\beta=1.0$ means the standard IoU loss. The best AP of AIoU is obtained when the squeezed ratio is 0.8.
\vspace{-0.4cm}
\subsection{Experiments on MS COCO}
\vspace{-0.2cm}
To further validate our method, we also evaluate our SCOD on MS COCO 2017. As shown in Table~\ref{sample-table6}. Without bells and whistles, when the input size is 320, our SCOD produces 34.0$\%$ mAP that is better than most of one-stage detectors. Our model outperforms the baseline DSSD320 by 6$\%$ AP (from 28.0$\%$ to 34.0$\%$). When the input size is increased to 512, our method can achieve 38.7$\%$ mAP with ResNet-101, which is comparable to most of state-of-the-art single-shot detectors at the same scale. Especially, our detector can achieve 40.5$\%$ mAP when we adopt the multi-scale inference strategy, which is competitive to most of one-stage detectors.
\vspace{-0.5cm}
\begin{table}[h]
  \vspace{-0.3cm}
  \caption{Results on MS COCO \texttt{test-dev} set. The $\dagger$ means multi-scale inference.}
  \footnotesize
  \label{sample-table6}
  \centering
  \resizebox{8.5cm}{2.2cm}{
  \begin{tabular}{p{2.3cm}|p{1.5cm}|p{0.4cm}p{0.35cm}p{0.35cm}p{0.35cm}p{0.35cm}p{0.35cm}}
    \toprule
    Method           &   Backbone      & AP & AP$_{50}$ & AP$_{75}$ &  AP$_{S}$& AP$_{M}$ &AP$_{L}$ \\

    \hline
    SSD300 \cite{liu2016ssd}             & VGG-16   &25.1 & 43.1&25.8& 6.6 & 25.9 & 41.4    \\
    DSSD321 \cite{fu2017dssd}            & ResNet-101       &28.0 & 46.1& 29.2&  7.4 & 28.1 & 47.6   \\
    RefineDet320 \cite{zhang2018single}  & ResNet-101& 32.0 &51.4& 34.2 &10.5& 34.7 &50.4                  \\
    DAFS320 \cite{li2019dynamic}        & ResNet-101& 33.2 &52.7 &35.7 &10.9& 35.1 &52.0  \\
    \textbf{SCOD320}(Ours)              & ResNet-101    & 34.0 & 53.0& 36.0&13.4 & 36.6 &50.2  \\
    \hline
    SSD512 \cite{liu2016ssd}           & VGG-16     & 28.8 &48.5& 30.3&10.9 & 31.8 & 43.5   \\
    DSSD513 \cite{fu2017dssd}            & ResNet-101       & 33.2 & 53.3& 35.2&13.0 & 35.4 & 51.1  \\
    RefineDet512 \cite{zhang2018single} &ResNet-101 &36.4 &57.5 &39.5 &16.6 &39.9 &51.4                    \\
    RetinaNet800 \cite{lin2017focal}            & ResNet-101     &39.1 &59.1 &42.3 &21.8 &42.7 &50.2  \\
    DAFS \cite{li2019dynamic}       & ResNet101 &38.6 &58.9 &42.2 &17.2 &42.2 &54.8   \\
    \textbf{SCOD512}(Ours)             & ResNet101    & 38.7 & 57.8&  42.9&20.3& 43.5 & 51.7      \\
    \textbf{SCOD512}$\dagger$ (Ours)           & ResNet101    & \textbf{40.5} &  59.9 &44.5 &22.5 &44.5& 53.2      \\
    \bottomrule
  \end{tabular}
  }
  \vspace{-0.4cm}
\end{table}

\vspace{-0.1cm}
\section{Conclusion}
\vspace{-0.4cm}
In this paper, we propose a scale-aware object detector (SCOD), in which two novel contributions are included. First, in SCOD, we introduce a scale-customized weak segmentation (SCWS) block to handle the problem of scale variation. Second, we introduce a new adaptive IoU (AIoU) loss to solve the imbalance problem between hard and easy samples by bounding box squeezing operator with a scale ratio.

\bibliographystyle{IEEEbib}
\bibliography{refs}

\begin{thebibliography}{10}

\bibitem{girshick2015fast}
Ross Girshick,
\newblock ``Fast r-cnn,''
\newblock in {\em Proceedings of the IEEE international conference on computer
  vision}, 2015, pp. 1440--1448.

\bibitem{ren2015faster}
Shaoqing Ren, Kaiming He, Ross Girshick, and Jian Sun,
\newblock ``Faster r-cnn: Towards real-time object detection with region
  proposal networks,''
\newblock in {\em Advances in neural information processing systems}, 2015, pp.
  91--99.

\bibitem{lin2017feature}
Tsung-Yi Lin, Piotr Doll{\'a}r, Ross Girshick, Kaiming He, Bharath Hariharan,
  and Serge Belongie,
\newblock ``Feature pyramid networks for object detection,''
\newblock in {\em Proceedings of the IEEE conference on computer vision and
  pattern recognition}, 2017, pp. 2117--2125.

\bibitem{dai2016r}
Jifeng Dai, Yi~Li, Kaiming He, and Jian Sun,
\newblock ``R-fcn: Object detection via region-based fully convolutional
  networks,''
\newblock in {\em Advances in neural information processing systems}, 2016, pp.
  379--387.

\bibitem{he2017mask}
Kaiming He, Georgia Gkioxari, Piotr Doll{\'a}r, and Ross Girshick,
\newblock ``Mask r-cnn,''
\newblock in {\em Proceedings of the IEEE international conference on computer
  vision}, 2017, pp. 2961--2969.

\bibitem{cai2018cascade}
Zhaowei Cai and Nuno Vasconcelos,
\newblock ``Cascade r-cnn: Delving into high quality object detection,''
\newblock in {\em Proceedings of the IEEE conference on computer vision and
  pattern recognition}, 2018, pp. 6154--6162.

\bibitem{Cao_2020_CVPR}
Yuhang Cao, Kai Chen, Chen~Change Loy, and Dahua Lin,
\newblock ``Prime sample attention in object detection,''
\newblock in {\em Proceedings of the IEEE/CVF Conference on Computer Vision and
  Pattern Recognition (CVPR)}, June 2020.

\bibitem{pang2019libra}
Jiangmiao Pang, Kai Chen, Jianping Shi, Huajun Feng, Wanli Ouyang, and Dahua
  Lin,
\newblock ``Libra r-cnn: Towards balanced learning for object detection,''
\newblock in {\em Proceedings of the IEEE Conference on Computer Vision and
  Pattern Recognition}, 2019, pp. 821--830.

\bibitem{li2019scale}
Yanghao Li, Yuntao Chen, Naiyan Wang, and Zhaoxiang Zhang,
\newblock ``Scale-aware trident networks for object detection,''
\newblock in {\em Proceedings of the IEEE International Conference on Computer
  Vision}, 2019, pp. 6054--6063.

\bibitem{liu2016ssd}
Wei Liu, Dragomir Anguelov, Dumitru Erhan, Christian Szegedy, Scott Reed,
  Cheng-Yang Fu, and Alexander~C Berg,
\newblock ``Ssd: Single shot multibox detector,''
\newblock in {\em European conference on computer vision}. Springer, 2016, pp.
  21--37.

\bibitem{fu2017dssd}
Cheng-Yang Fu, Wei Liu, Ananth Ranga, Ambrish Tyagi, and Alexander~C Berg,
\newblock ``Dssd: Deconvolutional single shot detector,''
\newblock {\em arXiv preprint arXiv:1701.06659}, 2017.

\bibitem{redmon2016you}
Joseph Redmon, Santosh Divvala, Ross Girshick, and Ali Farhadi,
\newblock ``You only look once: Unified, real-time object detection,''
\newblock in {\em Proceedings of the IEEE conference on computer vision and
  pattern recognition}, 2016, pp. 779--788.

\bibitem{lin2017focal}
Tsung-Yi Lin, Priya Goyal, Ross Girshick, Kaiming He, and Piotr Doll{\'a}r,
\newblock ``Focal loss for dense object detection,''
\newblock in {\em Proceedings of the IEEE international conference on computer
  vision}, 2017, pp. 2980--2988.

\bibitem{zhang2018single}
Shifeng Zhang, Longyin Wen, Xiao Bian, Zhen Lei, and Stan~Z Li,
\newblock ``Single-shot refinement neural network for object detection,''
\newblock in {\em Proceedings of the IEEE conference on computer vision and
  pattern recognition}, 2018, pp. 4203--4212.

\bibitem{zhao2019m2det}
Qijie Zhao, Tao Sheng, Yongtao Wang, Zhi Tang, Ying Chen, Ling Cai, and Haibin
  Ling,
\newblock ``M2det: A single-shot object detector based on multi-level feature
  pyramid network,''
\newblock in {\em Proceedings of the AAAI Conference on Artificial
  Intelligence}, 2019, vol.~33, pp. 9259--9266.

\bibitem{wang2020single}
Keyang Wang and Lei Zhang,
\newblock ``Single-shot two-pronged detector with rectified iou loss,''
\newblock in {\em Proceedings of the 28th ACM International Conference on
  Multimedia}, 2020, pp. 1311--1319.

\bibitem{wu2019iou}
Shengkai Wu and Xiaoping Li,
\newblock ``Iou-balanced loss functions for single-stage object detection,''
\newblock {\em arXiv preprint arXiv:1908.05641}, 2019.

\bibitem{Zhu_2019_CVPR}
Chenchen Zhu, Yihui He, and Marios Savvides,
\newblock ``Feature selective anchor-free module for single-shot object
  detection,''
\newblock in {\em Proceedings of the IEEE/CVF Conference on Computer Vision and
  Pattern Recognition (CVPR)}, June 2019.

\bibitem{yu2016unitbox}
Jiahui Yu, Yuning Jiang, Zhangyang Wang, Zhimin Cao, and Thomas Huang,
\newblock ``Unitbox: An advanced object detection network,''
\newblock in {\em Proceedings of the 24th ACM international conference on
  Multimedia}, 2016, pp. 516--520.

\bibitem{everingham2007pascal}
Mark Everingham and John Winn,
\newblock ``The pascal visual object classes challenge 2007 (voc2007)
  development kit,''
\newblock {\em University of Leeds, Tech. Rep}, 2007.

\bibitem{lin2014microsoft}
Tsung-Yi Lin, Michael Maire, Serge Belongie, James Hays, Pietro Perona, Deva
  Ramanan, Piotr Doll{\'a}r, and C~Lawrence Zitnick,
\newblock ``Microsoft coco: Common objects in context,''
\newblock in {\em European conference on computer vision}. Springer, 2014, pp.
  740--755.

\bibitem{li2019dynamic}
Shuai Li, Lingxiao Yang, Jianqiang Huang, Xian-Sheng Hua, and Lei Zhang,
\newblock ``Dynamic anchor feature selection for single-shot object
  detection,''
\newblock in {\em Proceedings of the IEEE International Conference on Computer
  Vision}, 2019, pp. 6609--6618.

\end{thebibliography}

\end{document}